\documentclass[10pt,twocolumn,letterpaper]{article}

\usepackage[pagenumbers,algorithms]{wacv}   % REVIEW version, Algorithms track review,
\usepackage{amsmath}
\usepackage{amssymb}
\usepackage{amsfonts}
\usepackage{booktabs}
\usepackage{tabularx}
\usepackage{multirow}
\usepackage{graphicx}
\usepackage{subcaption}
\usepackage{float}
\usepackage[dvipsnames,table]{xcolor}
\usepackage{xspace}

\usepackage[linesnumbered,ruled,vlined]{algorithm2e}

\graphicspath{{imgs/}{./}}

\definecolor{lilac}{rgb}{0.71, 0.4, 0.82}
\newcommand{\method}{\textcolor{lilac}{LILAC}\xspace}

\usepackage[globalcitecopy]{bibunits}        % NOU (după preamble)
\definecolor{wacvblue}{rgb}{0.21,0.49,0.74}
\usepackage[pagebackref,breaklinks,colorlinks,allcolors=wacvblue]{hyperref}

\usepackage[capitalize]{cleveref}

\def\wacvPaperID{471} % *** Enter the WACV Paper ID here (from CMT)
\def\confName{WACV}
\def\confYear{2027}

\title{\textcolor{lilac}{LILAC}: \textcolor{lilac}{L}ayer-Wise
\textcolor{lilac}{I}ndependent \textcolor{lilac}{L}oR\textcolor{lilac}{A}s and
\textcolor{lilac}{C}ascaded Conditioning for Multi-Concept Customization of
Diffusion Models}

\author{%
Marian Lupa\c{s}cu\textsuperscript{1,2}\quad
Sebastian Ripa\textsuperscript{1,3}\quad
Mihai Tr\u{a}sc\u{a}u\textsuperscript{1}\quad
Mariana-Iuliana Georgescu\textsuperscript{1}\quad
Ionu\c{t} Mironic\u{a}\textsuperscript{1}\\[2pt]
{\normalsize\textsuperscript{1}Adobe Research, Romania \quad
\textsuperscript{2}University of Bucharest, Romania \quad
\textsuperscript{3}International Computer High School of Bucharest, Romania}\\[2pt]
{\tt\small \{lupascu, mtrascau, mgeorgescu, mironica\}@adobe.com}
}

\begin{document}
\twocolumn[{%
\renewcommand\twocolumn[1][]{#1}%
\maketitle
% Teaser is generated by `python scripts/make_teaser.py` from the existing
% bench_baseline / bench_mvsd_anti2 outputs.

\vspace{-1cm}
\includegraphics[width=0.99\linewidth]{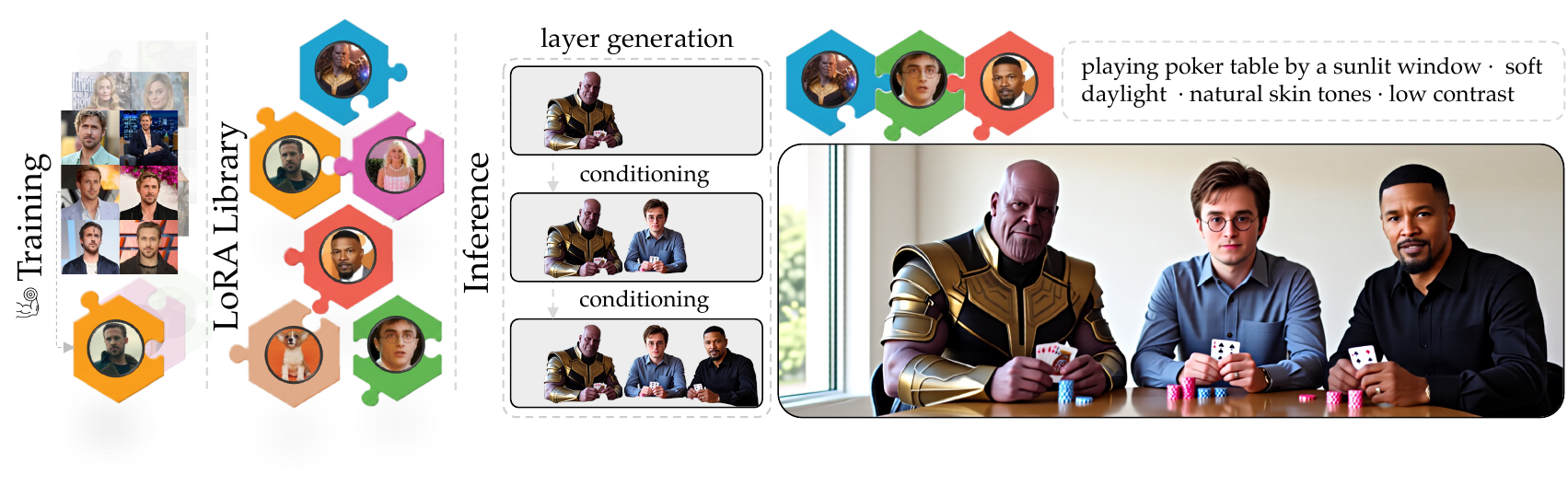}
\vspace{-1cm} 
\captionof{figure}{Overview of \method. \textbf{Training} (left): Each concept is trained in isolation into a single LoRA adapter, building a library of independently trained, per-concept adapters. 
\textbf{LoRA Library} (center): Concepts trained independently, one per subject.
\textbf{Inference} (right): \method selects a subset and renders every subject in one coherent scene from a text prompt. Exactly one adapter is active per pass under frozen conditioning, so identities never interfere at the parameter level, with no joint training and no weight merging. Each identity is
preserved; in this example three subjects share a poker table by a sunlit window.}\label{fig:teaser}
}
\vspace{0.4cm}
% \label{fig:teaser}
]

\begin{abstract}
Personalizing text-to-image diffusion models to render several specific subjects in a coherent image remains challenging: the model must preserve each subject's identity while keeping the scene spatially and visually coherent. Methods that fuse independently trained concept adapters in a shared weight space (via federated averaging, gradient fusion, or orthogonality constraints) suffer from identity confusion and style bleeding. In this work, we show that \textit{composing concepts as separate image layers}, instead of merging their adapters in a shared weight space, \textit{avoids parameter-level interference}. We introduce \method, a framework that composes independently trained low-rank adapters at inference time: each subject is conditioned on the frozen composite of previously placed subjects, with exactly one adapter active at a time, therefore identities never interfere at the parameter level. \method composes the adapters without joint training, scales linearly with the number of concepts, and is backbone-agnostic. Under the Orthogonal Adaptation protocol, \method applied on Qwen-Image-Edit+Qwen-Image-Layered reaches an ArcFace detection rate of $0.861$. Code is available at \url{https://github.com/marianlupascu/LILAC}.

% \href{https://anonymous.4open.science/r/LILAC-9820/README.md}{\texttt{anonym.link}}.

% \url{https://github.com/marianlupascu/LILAC}. 
% \href{https://anonymous.4open.science/r/LILAC-9820/README.md}{\texttt{Anonymous repository}}.

% TODO (review): anonymize this repository link before submission (double-blind).
  % \keywords{Multi-concept personalization \and Diffusion models \and
  % Low-rank adaptation \and Layered generation \and Identity preservation}
\end{abstract}

\section{Introduction}
\label{sec:intro}
% \vspace{-0.25cm}

Parameter-efficient adaptation has made single-concept personalization of text-to-image
diffusion models routine: a low-rank adapter (\eg LoRA~\cite{hu2022lora}) fine-tuned on a
handful of images reliably captures one subject. Rendering several specific subjects in one coherent image is harder: the model must keep each identity while arranging the subjects plausibly. The standard recipe trains one adapter per concept and then combines them in a shared parameter space. This is where the difficulty concentrates: once two concepts occupy the same weights their representations interfere, as identities blend, styles leak across subjects, and one concept often dominates another.

\noindent
Existing remedies for combining adapters each trade one cost for another. Gradient fusion,
as in Mix-of-Show~\cite{gu2023mixofshow}, recovers fidelity at the price of per-composition
optimization. Orthogonality constraints, as in Orthogonal Adaptation~\cite{po2024orthogonal},
enforce separation by retraining every concept under a shared scheme, and hold only up to
the available rank budget as the number of concepts grows. 
% Selection-based merging such as ZipLoRA~\cite{shah2024ziplora} and K-LoRA~\cite{ouyang2025klora} remains confined to the
% subject-and-style setting. 
None of these can reuse the many adapters already trained independently. %A separate line of work composes concepts spatially,
% through masks, bounding boxes, or layout tokens, but it requires explicit spatial
% supervision and still merges or regionally blends the underlying representations.

\noindent
To mitigate weight-space interference while reusing existing adapters, we introduce \method, which recasts multi-subject generation as a \emph{layered synthesis} problem rather than a weight-merging one. Each concept is an independently trained adapter, and subjects are composed one at a time at inference: placed sequentially, each pass conditioned on the frozen composite of the subjects already placed, with exactly one adapter active. Because no two adapters are ever active together, the cross-concept interference that orthogonality-based methods minimize never arises, and identities cannot interfere at the parameter level. Geometric and photometric coherence instead emerges from the conditioning itself, with no auxiliary objective, optimization loop, or spatial supervision. \method needs no joint training and no optimization at composition, scales linearly in the number of subjects, and composes adapters trained in isolation, as shown in Fig.~\ref{fig:teaser}.

\method is backbone agnostic, therefore, we implement it on top of LaDe (a layered RGBA diffusion model)~\cite{lungustan2026lade} and Qwen-Image-Edit+Qwen-Image-Layered~\cite{yin2025qwenlayered}. For non-layered generative methods (Qwen-Image-Edit~\cite{wu2025qwenimage}), we propose the \emph{scaffold} and \emph{decomposition} techniques to enable the application of \method.
We evaluate \method{} on the benchmark introduced by Orthogonal Adaptation~\cite{po2024orthogonal}, achieving an ArcFace detection rate of 0.861.

\noindent
Our main contributions are: (i) we recast multi-concept customization as a layered,
inference-time composition problem and introduce \method, a cascaded protocol that composes
independently trained adapters without ever merging their weights; 
(ii) a per-concept LoRA
binding that eliminates cross-concept interference at the parameter level by construction,
rather than minimizing it through a training-time orthogonality constraint; 
(iii) an empirical evaluation under the Orthogonal Adaptation protocol
in which our configuration reaches an ArcFace detection rate of $0.861$.
% and
% (iv) a controlled same-backbone analysis (\cref{sec:robustness}) showing that the baseline 
% weight-merging failure is robust to the baseline's own hyperparameters such as LoRA rank,
% merge scale, subject count, and the orthogonality constraint itself---and is
% therefore a property of weight-space fusion on the backbone rather than an untuned
% baseline.

% (iii) a
% procedure that needs no joint training, no per-composition optimization, and no constraint
% on how the adapters were trained, and that therefore composes off-the-shelf adapters and
% scales linearly in the number of subjects; 

% (iv) a backbone-agnostic formulation, validated
% under a single protocol on both an open-source image-editing model and a native layered
% diffusion model; and

\section{Related Work}
\label{sec:related}
% \vspace{-0.25cm}

\noindent
\textbf{Subject-driven personalization.}
DreamBooth~\cite{ruiz2023dreambooth} fine-tunes the full backbone around a
rare identifier token, Textual Inversion~\cite{gal2023textual} optimizes a new token
embedding with the model frozen, and LoRA~\cite{hu2022lora} injects low-rank
weight increments that capture a concept cheaply. Later work
refines what is adapted: B-LoRA~\cite{frenkel2024blora} separates content
and style by training only selected attention blocks.
These methods each produce one high-quality adapter per concept, and a large
ecosystem of such adapters now exists, but none address how to place several
independently learned concepts in one image, the problem we study.

\noindent
\textbf{Multi-concept customization by weight-space fusion.}
The prevailing approach combines per-concept adapters in a
shared parameter space. Custom Diffusion~\cite{kumari2023custom} jointly optimizes a small
set of weights over all concepts, while Mix-of-Show~\cite{gu2023mixofshow} tunes
an embedding-decomposed LoRA per concept and merges them by gradient fusion.
ZipLoRA~\cite{shah2024ziplora} learns coefficients that combine a subject and a
style adapter, and K-LoRA~\cite{ouyang2025klora} selects, training-free, the Top-$K$ most
important entries of each adapter per attention layer. Orthogonal
Adaptation~\cite{po2024orthogonal} constrains independently trained adapters to
orthogonal subspaces so their residuals sum with reduced interference.
All of these represent multiple concepts in one set of weights, so residuals
interact and identities can blend, and several also require joint training, a shared
constraint, or per-composition optimization; ZipLoRA and K-LoRA moreover target
subject-and-style pairing rather than multiple distinct subjects. Unlike these,
\method \textit{never forms a combined weight, so identities cannot interfere at the weight level}.

\noindent
\textbf{Spatial and layout control.}
A second line composes concepts by specifying where to place them:
Mix-of-Show~\cite{gu2023mixofshow} binds each concept to a spatial region through
regionally controllable sampling, GLIGEN~\cite{li2023gligen} grounds generation on
bounding boxes and layout tokens, grid-based LoRA~\cite{abdal2025gridlora} uses a
structured grid layout, and OMG~\cite{kong2024omg} targets occlusion between subjects.
These achieve strong control but depend on explicit spatial supervision (masks,
boxes, or layouts) supplied with the prompt. \method introduces no such supervision: placement, scale, and interaction follow from the generation order and the frozen context alone, which also carries consistent lighting, and concept separation is \emph{structural rather than spatial}.

\noindent
\textbf{Layered and transparent generation.}
A body of work generates images as explicit layers.
LayerDiffuse~\cite{zhang2024layerdiffuse} encodes alpha transparency as a latent offset,
turning a latent diffusion model into a generator of transparent RGBA layers
conditioned on a fixed background. Recent models recover an arbitrary number of
semantically disentangled RGBA layers from a single image~\cite{yin2025qwenlayered}, and
LaDe~\cite{lungustan2026lade} generates multiple distinct RGBA layers within one model,
conditioning each on those already produced. We build directly on this
capability, but such models target the generation of transparent assets, not the
identity-preserving composition of several independently trained concepts.
Our contribution is the \textit{cascaded, per-concept-bound protocol that turns single-layer
conditional generation into scalable multi-subject customization}.

\section{Method}
\label{sec:method}
% \vspace{-0.25cm}

\begin{figure}[t]
\centering
% \vspace{-0.1cm}
\includegraphics[width=\linewidth]{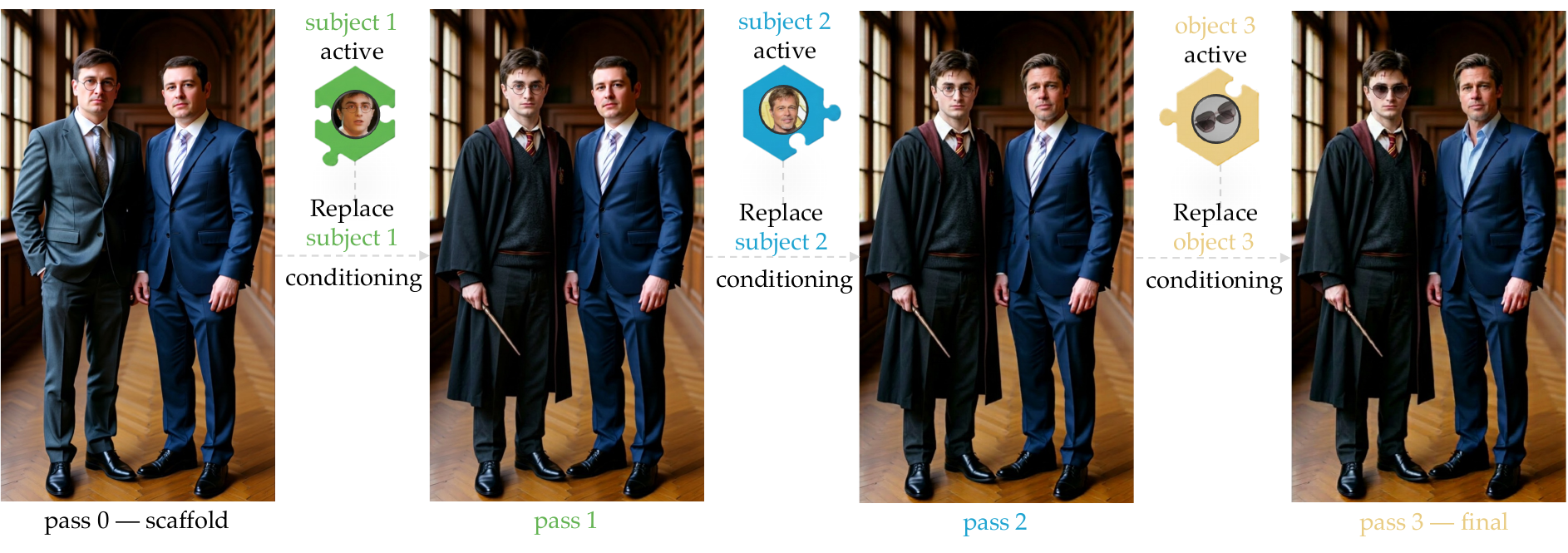}
\vspace{-0.5cm}
\caption{Cascaded inference (scaffold configuration). Given an image (natural or generated), each concept is then added in its own
pass with only its adapter active, conditioned on the frozen image of the concepts
placed so far. No two adapters are active together, and earlier concepts and the background
are preserved as later ones are added; the final pass adds a non-person concept
(sunglasses), showing that the procedure composes objects and attributes as well as
identities.}
% \vspace{-0.5cm}
\label{fig:pipeline}
\end{figure}

Our key observation is that \textbf{multi-subject generation} does not require concepts to be
combined in a shared weight space at all. We propose to treat it instead \textbf{as a layered
synthesis problem}: each concept is captured by an adapter, and subjects are
composed at inference time by rendering them on separate transparent layers and
conditioning each generation on the layers produced. 
This recasting removes the source of cross-concept interference that affects weight-space fusion while requiring no joint training and no optimization at composition time. \cref{fig:pipeline} details the cascaded inference procedure.

\begin{figure}[t]
\centering
% \vspace{-0.5cm}
\includegraphics[width=\linewidth]{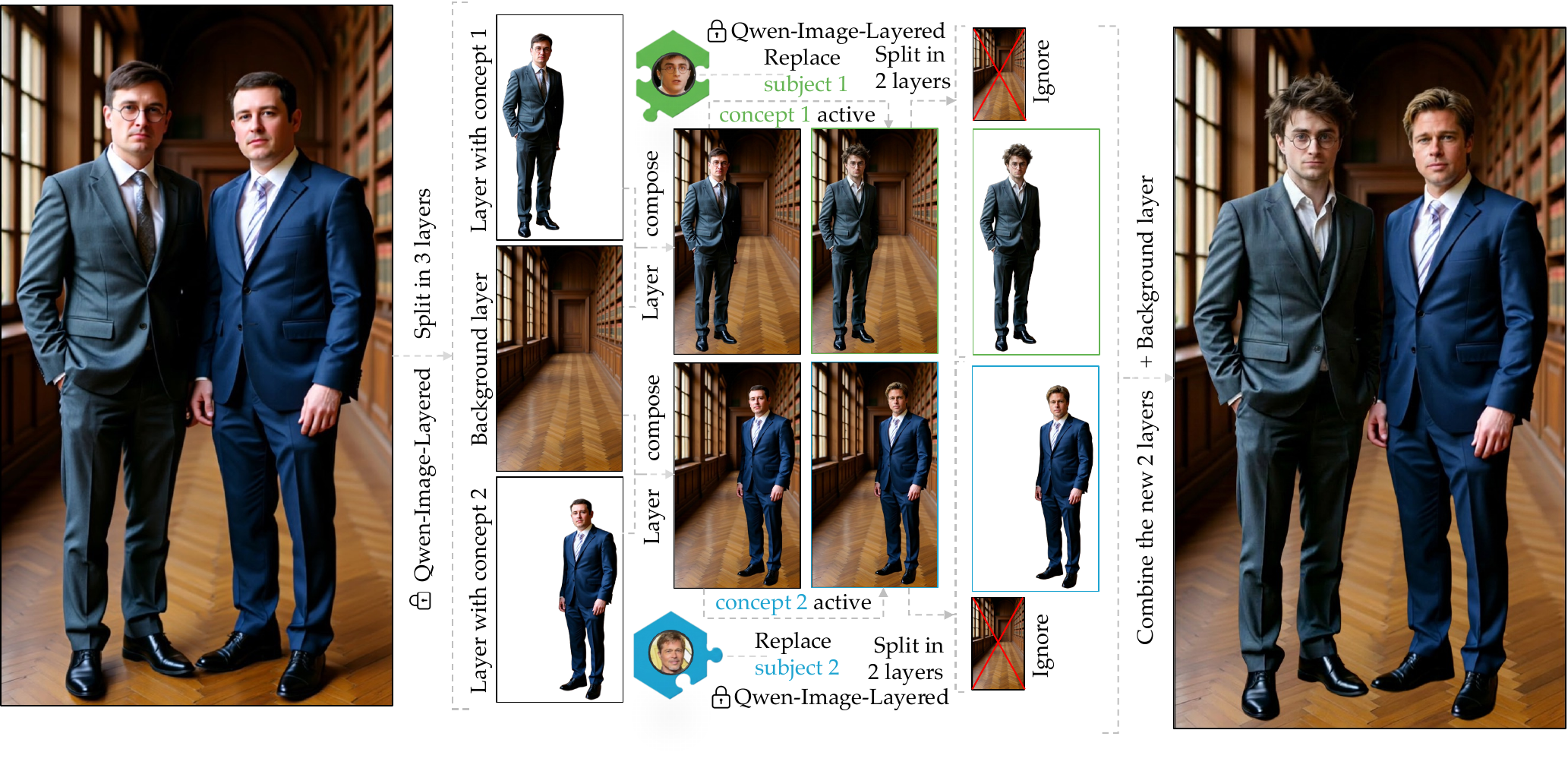}
\vspace{-0.75cm}
\caption{Decomposition configuration on Qwen-Image-Edit~\cite{wu2025qwenimage}. Given an input image, Qwen-Image-Layered~\cite{yin2025qwenlayered} then
splits it into RGBA layers, here a background and two subject layers. Each subject is
regenerated in its own pass: the frozen background is added as context, the placeholder
subject is replaced with that concept's adapter active, and the background is dropped to recover a transparent subject layer. The
background carries no adapter and stays unchanged throughout, then recombines with the two
regenerated subject layers into the final image. Because the per-subject layers come from a
layer split rather than a direct edit, this route inherits any error in the split, so it
preserves identity less reliably than the scaffold configuration (\cref{sec:comparison}).}
\label{fig:decomposition}

\vspace{-0.5cm}
\end{figure}

\subsection{Overview}
\label{sec:overview}
% \vspace{-0.1cm}

Given $N$ concepts, each described by a small set of reference images $X_k$, our goal is to synthesize a single image that contains all $N$ subjects with their individual
identities preserved and their interactions natural, using only inference-time
computation. Our method comprises three components: an independent adapter is trained for
each concept (\cref{sec:training}); the concepts are assigned to ordered layers
(\cref{sec:assignment}); and a cascaded inference procedure composes them on transparent
layers (or images) under frozen conditioning (\cref{sec:cascade}). We analyze the resulting isolation
property in \cref{sec:binding}.

\noindent
The method is defined over a layered diffusion backbone.
Formally, we assume a backbone $\mathfrak{B} = \langle \theta, E, D \rangle$ that
exposes three operations: an encoder $E$ and decoder $D$ for transparent RGBA layers; a
conditional sampler $\mathrm{Sample}(\cdot)$ that generates a layer from noise and a text
prompt, optionally conditioned on a clean layer or composite; and weights $\theta$ that
admit low-rank adaptation. 
% We make no assumption about the backbone's architecture or its
% prediction parameterization, whether $\epsilon$-, $v$-, or flow-based, since the sampler
% encapsulates these details. The two capabilities our method actually requires, RGBA output
% and conditioning on an existing layer, are provided natively by a backbone trained for
% layered generation, and can be approximated on a standard image model, either by editing
% each subject into a rendered scene or by decomposing that scene into RGBA layers with an
% off-the-shelf decomposer.
% We instantiate
% $\mathfrak{B}$ with an open-source image backbone (in two configurations) and a native
% layered backbone in our experiments, treating that choice as an
% implementation detail (\cref{sec:setup}).

% \noindent
% Throughout, we index the concepts by their generation order, so that $c_1$ is the anchor
% and $c_k$ is produced at pass $k$; this ordering is the assignment of
% \cref{sec:assignment}. We reserve $i$ and $j$ for two distinct concepts, write $\alpha_k$
% for the alpha channel of layer $L_k$, and denote alpha-over compositing by $\oplus$.

% \noindent
% It is important to separate what the backbone supplies from what our method contributes.
% The backbone provides the generative prior, the transparency model, and the ability to
% condition one generation on a fixed context; the geometric and photometric coherence
% between a new subject and the existing scene is a property of this pretrained
% conditioning, not something we introduce. 

\noindent
\method turns this single-layer conditional capability into identity-preserving multi-subject
generation: per-concept binding keeps each identity governed by its own adapter, and
the cascade composes arbitrarily many subjects from independently trained
adapters without merging them.

% In contrast to region- and layout-based multi-concept methods,
% which steer placement through explicit spatial masks, bounding boxes, or layout tokens,
% our method introduces no spatial supervision: composition is governed entirely by the
% generation order and the frozen context, and the separation of concepts is structural
% rather than spatial.

\subsection{Independent Per-Concept LoRA Training}
\label{sec:training}
% \vspace{-0.1cm}

For each concept $c_k$ we train a dedicated low-rank adapter on its reference set $X_k$
while keeping the backbone frozen. The adapter modifies a pretrained weight
$W_0 \in \mathbb{R}^{d \times d}$ through a low-rank increment,
\begin{equation}
W_0 + \Delta W_k, \qquad \Delta W_k = B_k A_k,
\label{eq:lora}
\end{equation}
with $B_k \in \mathbb{R}^{d \times r}$, $A_k \in \mathbb{R}^{r \times d}$, and rank
$r \ll d$, so that the per-concept storage and the adaptation cost remain small. We
optimize only the adapter parameters by minimizing the backbone's native generative
objective $\ell_{\mathfrak{B}}$ on the concept's images $X_k$,
\begin{equation}
\min_{B_k,\, A_k} \;\; \mathbb{E}_{x \sim X_k}\big[\, \ell_{\mathfrak{B}}(x;\, \theta,\,
\Delta W_k,\, p_k) \,\big],
\label{eq:train}
\end{equation}
where $\theta$ are the parameters of the base model, $\Delta W_k$ are the adapter weights,  and $p_k$ is a caption that binds the concept $c_k$ to a unique identifier token. %, used only to
% keep the token of one concept from colliding with that of another. % Because
% $\ell_{\mathfrak{B}}$ is the objective the backbone is already trained with, \cref{eq:train}
% inherits its parameterization without modification.

\noindent
This stage is performed independently for every
concept: there is no joint optimization, no shared basis, and no orthogonality or
separation constraint coupling the adapters. This decoupling is what later lets the cascade combine arbitrary
adapters at inference while keeping each identity governed solely by its own (\cref{sec:binding}).

% As a consequence, an adapter is nothing more than a standard single-concept LoRA, and our method
% can equally compose adapters that were trained in isolation by others, without retraining
% or adaptation. 

\subsection{Layer Assignment and Anchor Selection}
\label{sec:assignment}
% \vspace{-0.1cm}

Before the cascade (\cref{sec:cascade}), the concepts are ordered: an
assignment maps each concept to a layer index in $\{1,\dots,N\}$, with index $1$ the anchor,
so $c_1$ is generated first and $c_k$ at step $k$. Unlike the adapters, this assignment is a design choice rather than a learned component and is fixed before inference.

\noindent
Because each subject is generated conditioned on the previous ones, earlier concepts shape the placement, scale,
and interaction of later ones, and the anchor fixes the global composition.
The order usually follows from the prompt or a user specification; we treat it as a hyperparameter and find in the supplementary material that composition order has only a minor effect.

% \noindent
% Whatever rule is used, the assignment scales linearly with the number of concepts: it
% allocates $N$ ordered slots and is computed once, in advance, with no combinatorial
% search over concept combinations. This keeps the overall procedure linear in $N$ and free
% of the assignment-time blow-up that affects methods enforcing pairwise constraints among
% concepts.

\subsection{Cascaded Inference with Frozen Layer Cond.}
\label{sec:cascade}
% \vspace{-0.1cm}

The core of our method composes subjects additively in layer space rather than weight
space, with no retraining or joint optimization. Given the per-concept adapters
$\{\Delta W_k\}_{k=1}^{N}$ from \cref{sec:training}, we generate subjects one at a time:
each is rendered on an RGBA layer (or image, as in~\cref{fig:pipeline}), then frozen and
reused as fixed conditioning context for every later pass. This frozen context enforces
geometric and photometric consistency across subjects while keeping each
identity governed by a single adapter. Throughout,
$\epsilon_{(k)} := \epsilon_{\theta_0 + \Delta W_k}$ denotes the backbone with only the
$k$-th increment applied, whose weight-level isolation we analyse in \cref{sec:binding}.

\noindent
% We present the cascade in its native layered form here, and describe the two configurations
% we use on a standard image backbone, scaffold and decomposition, at the end of the section. 
We begin with the anchor (layer $k=1$, \cref{sec:assignment}). It is generated first and
without conditioning, establishing the global composition, pose,
scale, and illumination to which the remaining subjects conform:
\begin{equation}
  \hat{z}_1 = \mathrm{Sample}\!\left(\epsilon_{(1)};\; x_T,\, p_1,\, \varnothing\right),
  \qquad
  L_1 = D(\hat{z}_1),
  \label{eq:anchor}
\end{equation}

\begin{algorithm}[t]
\SetAlgoLined
\DontPrintSemicolon
\caption{Cascaded inference with frozen layer conditioning}
\label{alg:cascade}
\KwIn{adapters $\{\Delta W_k\}_{k=1}^{N}$ in generation order; prompts $\{p_k\}$;
      frozen backbone $\mathfrak{B}=\langle \epsilon_\theta, E, D\rangle$}
\KwOut{multi-subject image $I$; per-subject layers $\{L_k\}_{k=1}^{N}$}
$\mathcal{C}_0 \leftarrow \varnothing$ \tcp*{empty (transparent) canvas}
\For{$k \leftarrow 1$ \KwTo $N$}{
  $x_T \sim \mathcal{N}(0,\mathbf{I})$ \tcp*{fresh noise}
  $z^{c} \leftarrow E(\mathcal{C}_{k-1})$ \tcp*{clean scaffold latent ($\varnothing$ if $k{=}1$)}
  $\epsilon_{(k)} \leftarrow \epsilon_{\theta_0 + \Delta W_k}$ \tcp*{bind only concept $c_k$}
  $\hat{z}_k \leftarrow \mathrm{Sample}(\epsilon_{(k)};\, x_T,\, p_k,\, z^{c})$\;
  $L_k \leftarrow D(\hat{z}_k)$ \tcp*{new RGBA layer}
  $\mathcal{C}_k \leftarrow L_k \oplus \mathcal{C}_{k-1}$ \tcp*{freeze \& accumulate}
}
$I \leftarrow \mathcal{C}_N$\;
\Return $I,\ \{L_k\}_{k=1}^{N}$\;
\end{algorithm}

\noindent
where $x_T \sim \mathcal{N}(0,\mathbf{I})$ is the initial latent and $p_1$ the anchor
prompt. The resulting layer $L_1$ carries an alpha matte $\alpha_1$ that delimits the
anchor's spatial support and leaves the remainder of the canvas transparent.

\noindent
Once produced, a layer is frozen: its pixels and alpha matte are never
revisited. To condition the next pass, we alpha-composite all frozen layers
and encode the result into a clean, noise-free latent,
\begin{equation}
  \mathcal{C}_{k-1} = L_{k-1} \oplus \mathcal{C}_{k-2},
  \quad \mathcal{C}_0 = \varnothing,
  \qquad
  z^{c}_{k-1} = E\!\left(\mathcal{C}_{k-1}\right),
  \label{eq:composite}
\end{equation}
where $\oplus$ denotes the alpha-over operator and $\mathcal{C}_0$ is an empty canvas.
Because $z^{c}_{k-1}$ is supplied at the clean ($t=0$) end, it acts as fixed conditioning context and is not denoised.
% The accumulated layers thus act as a structural scaffold for the generation that follows, ensuring that later subjects are placed coherently with respect to the earlier ones.

\noindent
The clean latent $z^{c}_{k-1}$ is supplied through the
backbone's reference-conditioning channel; since we use two different layered backbones, their precise conditioning is given in \cref{sec:setup}.

% it is appended to the noisy foreground latent
% along the token dimension, and the backbone's attention layers attend jointly over the
% two, so that the foreground is denoised while reading position, scale, and appearance
% from the fixed context. Since $z^{c}_{k-1}$ is the clean encoding of the composite and is
% held at the $t=0$ level throughout sampling, it provides keys and values to the foreground
% tokens but is never itself updated; this is the mechanism by which the scaffold shapes the
% new layer without being altered by it. T
\noindent
Each remaining concept $c_k$ ($k \in \{2,\dots,N\}$) is then synthesised in a new
pass (\cref{fig:pipeline}) that activates only $\Delta W_k$ and conditions on the previously generated image:

\vspace{-0.3cm}
\begin{equation}
\begin{split}
  \hat{z}_k = \mathrm{Sample}\!\left(\epsilon_{(k)};\; x_T,\, p_k,\, z^{c}_{k-1}\right), \\
  \quad
  L_k = D(\hat{z}_k), 
  \quad
  \mathcal{C}_k = L_k \oplus \mathcal{C}_{k-1},
  \label{eq:subsequent}
\end{split}
\end{equation}

\noindent
with $x_T$ redrawn at every pass. 

\noindent
Since $z^{c}_{k-1}$ encodes the position,
scale, pose, and lighting of all subjects placed so far, geometric and photometric coherence emerge from the conditioning itself, with no
auxiliary objective.
%; since $L_k$ is produced against $\mathcal{C}_{k-1}$, its alpha matte
% already reflects the occlusion with earlier subjects, and the alpha-over composite resolves
% any residual overlap. 
% and through the backbone's shared
% attention the new subject is generated to match this context, adopting a compatible scale
% and illumination and interaction-appropriate cues (gaze, body orientation, proximity)
% without any explicit spatial constraint or layout token in $p_k$. Because frozen layers
% cannot be modified by a later pass, ordered occlusion is handled naturally, while close
% mutual occlusion and physical contact remain the hardest cases (\cref{sec:limitations}).

\noindent
After all $N$ layers are generated, the final image is their alpha composite,
\begin{equation}
  I = \mathcal{C}_N = L_N \oplus L_{N-1} \oplus \cdots \oplus L_1 .
  \label{eq:final}
\end{equation}
% When the backbone emits true RGBA layers, every subject occupies a dedicated
% alpha-bounded layer, so its pixels remain physically separated in the final result and
% individual subjects can be reordered, edited, or replaced after generation without
% disturbing the others. This post-hoc controllability is unavailable to weight-space
% fusion, where all subjects share a single rasterized output and a single set of parameters.

\noindent
The procedure integrates with any layered backbone $\mathfrak{B}$ exposing
the operations of \cref{sec:overview}: \method leaves the backbone frozen, retrains no
adapter, and takes no gradient step at composition. It relies only on these abstract
operations, so on a backbone without native transparency the frozen prior is supplied
through an image-editing channel rather than an intact RGBA layer, with the per-concept
binding unchanged (\cref{sec:setup}). Its only added cost is $N$ sequential passes, linear in the number of subjects. \cref{alg:cascade} summarizes the routine.
% We emphasize that this cost is incurred entirely at sampling time: the method performs no
% optimization, no weight merging, and no retraining when concepts are composed, unlike
% fusion approaches that require per-composition gradient updates or a separate optimization
% stage. The price of avoiding these is additional sampling time, which we report in
% \cref{sec:setup}; 

\noindent
A backbone trained for layered generation performs this protocol directly. A standard image model cannot
generate a subject as a conditional transparent layer, so we implement the same cascade in one
of two ways.

\noindent
In the \emph{scaffold} configuration, the first pass generates a complete scene populated with generic placeholder figures in the intended arrangement (or a natural image can be used instead), fixing every subject's placement at once, and each concept pass then \emph{replaces} one placeholder with
its subject through an image-editing pass that takes the frozen scene as its conditioning
context and activates only that concept's adapter. This approach is illustrated in~\cref{fig:pipeline}.

\noindent
In the \emph{decomposition}
configuration, given a generated or natural image, an off-the-shelf decomposer
(Qwen-Image-Layered~\cite{yin2025qwenlayered}) splits it into RGBA layers, and each person
layer is then re-rendered with its concept adapter and alpha-composited back. The  \emph{decomposition} operation is illustrated in~\cref{fig:decomposition}.

\noindent
Both share the per-concept binding and frozen conditioning of \cref{eq:subsequent} and differ only in how
the layers are obtained. Because the scaffold commits to all placements before any identity
is inserted and edits the rendered scene directly, it preserves identities more reliably,
whereas the decomposition route inherits any error made by the layer split. We therefore
adopt the scaffold configuration as our default and report the decomposition realization for
comparison (\cref{sec:comparison}). 

\subsection{Per-Layer LoRA Binding}
\label{sec:binding}
% \vspace{-0.1cm}

Each subject's identity is governed by
a single adapter that never shares the active parameter space with another. During the
pass that generates $c_k$, the backbone applies exactly one increment,
$\epsilon_{(k)} = \epsilon_{\theta_0 + \Delta W_k}$, so the adapted weights
are $W_0 + \Delta W_k$ and no other concept contributes. We call
this per-layer LoRA binding, and show it removes cross-concept interference
at the parameter level structurally, not through a training-time constraint.

Weight-space fusion methods take the opposite route. To place several concepts in one
model they form a single merged weight,
\begin{equation}
\underbrace{W_0 + \sum_{k=1}^{N}\Delta W_k}_{\text{weight-space fusion}}
\qquad\text{vs.}\qquad
\underbrace{W_0 + \Delta W_k}_{\text{ours, pass } k},
\label{eq:fusion-vs-ours}
\end{equation}
in which the residuals of distinct concepts act jointly on the same activations. Prior
work quantifies the resulting interference through
$\chi_{ij} = \lVert \Delta W_i^{\top}\Delta W_j \rVert_F$ between residuals $i \neq j$,
and preserves identity by driving $\chi_{ij}$ toward zero. Orthogonal
Adaptation~\cite{po2024orthogonal}, for instance, constrains the adapters to a shared
orthogonal basis so that $\Delta W_i^{\top}\Delta W_j \approx 0$. Such constraints can
only approximate the ideal, and their effectiveness is bounded by the available rank
budget as the number of merged concepts grows. Independently trained adapters are, in
fact, not orthogonal: across the person concepts of \cref{sec:setup}, the mean pairwise
overlap between residual subspaces is $0.106$ (see the supplementary material; normalised to
$[0,1]$, with $0$ denoting orthogonality), $5.3\times$ the $0.020$ chance level measured
for random adapters of the same per-layer dimensionality, so $\chi_{ij}$ is non-negligible
and orthogonality must be actively imposed rather than assumed.

\noindent
Per-layer binding avoids this trade-off. Because the merged sum on the left of
\cref{eq:fusion-vs-ours} is never formed, no two adapters are ever simultaneously active,
so the interference $\chi_{ij}$ is eliminated rather than minimized: it never arises.
The failure modes of fused models (identity confusion and style bleeding) are thus removed
structurally, regardless of how the adapters were trained.

\noindent
This is a parameter-level guarantee, not a claim of lossless output: on a native layered
backbone the intact RGBA layers carry it through to the final image, whereas on a
non-layered backbone each pass re-encodes the composite, a separate and much smaller source
of degradation (\cref{sec:comparison,sec:limitations}).

\section{Experiments}
\label{sec:experiments}
% \vspace{-0.25cm}

\subsection{Experimental Setup}
\label{sec:setup}
% \vspace{-0.1cm}

\paragraph{Backbones.}
Because our method only assumes a backbone that can condition a generation on a fixed
prior and apply a concept adapter (\cref{sec:overview}), we validate it on two backbones
of different kinds. The first is an open-source configuration built on Qwen-Image and
Qwen-Image-Edit~\cite{wu2025qwenimage}. This backbone does not generate transparent layers
conditionally, so we evaluate both image scaffold and decomposition configurations presented in \cref{sec:cascade}. 
In the image-scaffold configuration, given a  placeholder scene (generated or natural), each subject is then inserted by a
Qwen-Image-Edit pass that takes the frozen scene as its source image, with that concept's
adapter as the only one active, as illustrated in~\cref{fig:pipeline}.
In the decomposition configuration, the scene is
split into RGBA layers by Qwen-Image-Layered~\cite{yin2025qwenlayered}, and each person layer
is re-rendered with its adapter and recomposited, as illustrated in~\cref{fig:decomposition}.
Each adapter is trained with a blank source
image (\cref{sec:training}), which forces identity into the text-conditioned pathway, so at
inference the source channel carries only the existing scene while the adapter contributes
the new identity through its trigger token. 

The second backbone is LaDe~\cite{lungustan2026lade}, a
native layered backbone that generates every subject directly on a true RGBA layer, with a
single adapter active per layer, and conditions on the intact frozen composed layers. 

% Both backbones realize the same
% cascade and the same per-concept binding: the open-source configuration makes our results
% fully reproducible, while the layered backbone additionally yields editable per-subject
% RGBA layers. 
Validating the identical protocol on a plain image-editing model and on a
native layered model isolates the contribution of the cascade and the binding from that of
any single backbone, and shows that the protocol transfers even to a backbone with no
native notion of layers.

\paragraph{Concepts and evaluation.}
We adopt the evaluation protocol and the concept bank of Orthogonal
Adaptation~\cite{po2024orthogonal}. % , which lets us compare against weight-space fusion on a
% published benchmark rather than a setup of our own. 
Each concept is defined by a small set
of reference images and trained into an independent adapter (\cref{sec:training}); similarly to Orthogonal
Adaptation~\cite{po2024orthogonal}, we form
multi-subject groups by drawing concepts at random and render each group as a landscape
($3{:}1$) scene, which places subjects predominantly side by side and thus under-samples the
close-contact regime that is hardest for \method (\cref{sec:limitations}). We release the
multi-subject groups used in our experiments for further comparison. We report every metric over a fixed set
of randomly formed groups and seeds per configuration. Before composition we screen each
concept for single-concept fidelity and exclude any that fails, which separates
multi-subject interference from single-concept training failures (see the supplementary material).
% TODO: state the number of concepts, the group size, the number of groups, and seeds.

\paragraph{Metrics.}
We measure three complementary quantities, following the protocol of Orthogonal
Adaptation~\cite{po2024orthogonal}. \emph{Identity preservation} (ID) is the
ArcFace~\cite{deng2019arcface} detection rate: for each subject we take the highest cosine
similarity between its detected face and its reference faces, count the subject as
preserved when this exceeds the threshold of~\cite{po2024orthogonal} (cosine $>0.32$, i.e.\
ArcFace distance $<0.68$), and report the fraction of subjects preserved. This matches the
identity-alignment metric of~\cite{po2024orthogonal}, which records an ArcFace detection
below the same $0.68$ distance and reports the detection probability rather than the raw
distance. Because this rate applies a hard threshold, in \cref{sec:robustness} we
additionally report the raw ArcFace cosine before thresholding, which distinguishes
degraded identity from absent identity.
\emph{Concept fidelity} (IA) is the CLIP-I~\cite{radford2021clip} cosine similarity between
each subject's cropped region and its reference images, averaged over subjects.
\emph{Text alignment} (TA) is a CLIPScore of the composed image against the joint prompt:
$\max(0,\cos)\times 2.5$ on the CLIP-T~\cite{radford2021clip} cosine
(CLIPScore~\cite{hessel2021clipscore}), the convention also used by~\cite{po2024orthogonal}.
Reporting identity and concept fidelity per subject, rather than only over the full image,
lets us detect the identity blending to which weight-space fusion is prone.
% TODO: add DINO and/or VLM-as-judge if used

\paragraph{Baselines.}
Our quantitative comparison covers the multi-concept methods reported by Orthogonal
Adaptation~\cite{po2024orthogonal} on this exact concept bank and protocol: extended textual
conditioning (P+~\cite{voynov2023pplus}), joint fine-tuning (Custom
Diffusion~\cite{kumari2023custom}), DreamBooth-LoRA merged by federated averaging
(DB-LoRA~\cite{ruiz2023dreambooth,hu2022lora}), gradient-based fusion
(Mix-of-Show~\cite{gu2023mixofshow}), and orthogonality-constrained training (Orthogonal
Adaptation~\cite{po2024orthogonal}). % Other multi-concept approaches, grid-based composition
% (grid-based LoRA~\cite{abdal2025gridlora}) and training-free selection for the
% subject-and-style pairing (K-LoRA~\cite{ouyang2025klora}, ZipLoRA~\cite{shah2024ziplora}), are
% covered as related work (\cref{sec:related}) and are not reported under this protocol, so we
% do not include them in the quantitative comparison.
% TODO (Tier 4): ZipLoRA / K-LoRA / Grid-LoRA are not in OrthA Tab. 2, so they have no
% directly comparable numbers under this protocol --- either run them or keep them as
% qualitative/related-work mentions (decision pending).

\paragraph{Implementation details.}
The full adapter, optimization, sampling, and model configuration is given in the supplementary material.

\subsection{Multi-Subject Comparison}
\label{sec:comparison}
% \vspace{-0.1cm}

\cref{tab:comparison} reports \method against the multi-concept baselines under the concept
bank, metrics, and protocol detailed in the supplementary material; baseline numbers are quoted from the SDXL
results of~\cite{po2024orthogonal}, so absolute values are comparable in scale but indicative
across backbones (\cref{sec:limitations}).

Two patterns stand out. First, on Identity the merge-based baselines either collapse at
composition (DB-LoRA drops to $0.098$, $\Delta\,{-}0.585$) or, at best, hold
flat (Orthogonal Adaptation, $0.745$, $\Delta\,{+}0.005$), whereas \method reaches $0.861$
(scaffold) and $0.877$ (layered), the two highest scores in the
table. Second, the composition gap $\Delta$ separates the failure modes from ours: a large
negative $\Delta$ signals identity loss at merge time, while \method on the layered backbone
stays flat ($\Delta\,{-}0.002$), confirming that adding a subject adds a pass, not
a cross term (\cref{sec:binding}). Text and image fidelity remain comparable to the strongest
baseline throughout. Absolute identity is also lifted by the
stronger single-concept backbone ($0.961$ vs.\ $0.740$ for SDXL), so the cross-backbone comparison should be read together with
$\Delta$ rather than from the headline number alone.

Crucially, this holds on the same backbone. Run on Qwen-Image-Edit, the backbone \method uses, Orthogonal Adaptation collapses at three subjects to an ArcFace detection rate of $0.000$ (ID $\Delta\,{-}0.929$), versus $0.861$ for \method scaffold. The failure is specific to identity: image alignment even rises (IA $0.869$). We attribute this to the orthogonal increments acting jointly over the full token sequence rather than disjoint spatial regions, yielding plausible but blended faces: CLIP-based IA still scores them as the right kind of person while ArcFace rejects each as a biometric match. This is the parameter-level interference \method removes (\cref{sec:binding}); \cref{sec:robustness} characterises it as a threshold effect (raw ArcFace cosine $0.29$, identity degraded, not absent) that no OA rank, merge scale, or orthogonality setting recovers and that layer-wise composition of the same adapters avoids.

The residual single-to-multi gap of \method on Qwen-Image-Edit ($\Delta\,{-}0.100$ scaffold) is a different effect: not the parameter-level interference of merging, which is eliminated on both backbones, but the re-encoding drift of a non-layered backbone (\cref{sec:limitations}). It is an order of magnitude below the merge-induced loss of same-backbone Orthogonal Adaptation ($\Delta\,{-}0.929$), and it vanishes on LaDe ($\Delta\,{-}0.002$), where the layers stay intact---the mechanism behaving exactly as predicted.

\begin{table*}[t]
\centering
\caption{Multi-subject comparison on the Orthogonal Adaptation concept bank and protocol~\cite{po2024orthogonal}, for each metric we give the single-concept score (\emph{Single}), the multi-concept score (\emph{Multi}), and their difference ($\Delta$). For the weight-merging baselines, \emph{Multi} is the post-merge result and $\Delta$ is the merge-induced change; \method performs no weight merging (merge time \emph{none}), so its $\Delta$ is the single-concept-to-multi-subject composition gap rather than a merge cost. \method is reported on Qwen-Image-Edit~\cite{wu2025qwenimage} (decomposition, scaffold) and on a native layered backbone (LaDe~\cite{lungustan2026lade}), so absolute values are indicative across backbones. Top-block baseline numbers are quoted directly from Table~2 of~\cite{po2024orthogonal} (SDXL). \textbf{We additionally re-run Orthogonal Adaptation on Qwen-Image-Edit, the same backbone as \method, for a controlled same-backbone comparison.} Within each backbone, the best \emph{Multi} per metric among \method configurations is in bold. $^{\dagger}$The Qwen OA rate of $0.000$ reflects the hard $0.68$ ArcFace-distance threshold; \cref{sec:robustness} gives a threshold-free, mean-over-seeds characterization (raw ArcFace cosine $0.29$, no single-concept screening) that no rank, merge scale, or orthogonality setting recovers.}
% \vspace{-0.25cm}

\label{tab:comparison}
\setlength{\tabcolsep}{4pt}
\resizebox{\textwidth}{!}{%
\begin{tabular}{l c ccc ccc ccc}
\toprule
& & \multicolumn{3}{c}{Text $\uparrow$} & \multicolumn{3}{c}{Image $\uparrow$} & \multicolumn{3}{c}{Identity $\uparrow$} \\
\cmidrule(lr){3-5}\cmidrule(lr){6-8}\cmidrule(lr){9-11}
Method & Merge time & Single & Multi & $\Delta$ & Single & Multi & $\Delta$ & Single & Multi & $\Delta$ \\
\midrule
P+~\cite{voynov2023pplus}                          & $<$1\,s     & 0.643 & 0.643 & ---      & 0.683 & 0.683 & ---      & 0.515 & 0.515 & ---      \\
Custom Diffusion~\cite{kumari2023custom}           & $\sim$2\,s  & 0.668 & 0.673 & $+$0.005 & 0.648 & 0.623 & $-$0.025 & 0.504 & 0.408 & $-$0.096 \\
DB-LoRA (FedAvg)~\cite{ruiz2023dreambooth,hu2022lora} & $<$1\,s  & 0.613 & 0.682 & $+$0.069 & 0.744 & 0.531 & $-$0.213 & 0.683 & 0.098 & $-$0.585 \\
Mix-of-Show (FedAvg)~\cite{gu2023mixofshow}        & $<$1\,s     & 0.625 & 0.621 & $-$0.004 & 0.745 & 0.735 & $-$0.010 & 0.728 & 0.706 & $-$0.022 \\
Mix-of-Show (Grad.\ Fusion)~\cite{gu2023mixofshow} & $\sim$15\,m & 0.625 & 0.631 & $+$0.006 & 0.745 & 0.729 & $-$0.016 & 0.728 & 0.717 & $-$0.011 \\
Orthogonal Adaptation (SDXL)~\cite{po2024orthogonal}      & $<$1\,s     & 0.624 & 0.644 & $-$0.010 & 0.748 & 0.741 & $-$0.007 & 0.740 & 0.745 & $+$0.005 \\
\midrule
Orthogonal Adaptation (Qwen)~\cite{po2024orthogonal} & $<$1\,s & 0.750 & 0.664 & $-$0.086 & 0.750 & \textbf{0.869} & $+$0.119 & 0.929 & 0.000$^{\dagger}$ & $-$0.929 \\

\method (decomposition)                              & none        & 0.754 & \textbf{0.734} & $-$0.020 & 0.746 & 0.743 & $-$0.003 & 0.961 & 0.639 & $-$0.322 \\
\method (scaffold)                                   & none        & 0.754 & 0.711 & $-$0.043 & 0.746 & 0.738 & $-$0.008 & 0.961 & \textbf{0.861} & $-$0.100 \\
\midrule
\method (LaDe)                                        & none        & 0.759 & \textbf{0.758} & $-$0.001 & 0.761 & \textbf{0.745} & $-$0.016 & 0.879 & \textbf{0.877} & $-$0.002 \\
\bottomrule
\end{tabular}%
}
\end{table*}

\begin{figure*}[t]
\centering
% \vspace{-0.1cm}
\includegraphics[width=\linewidth]{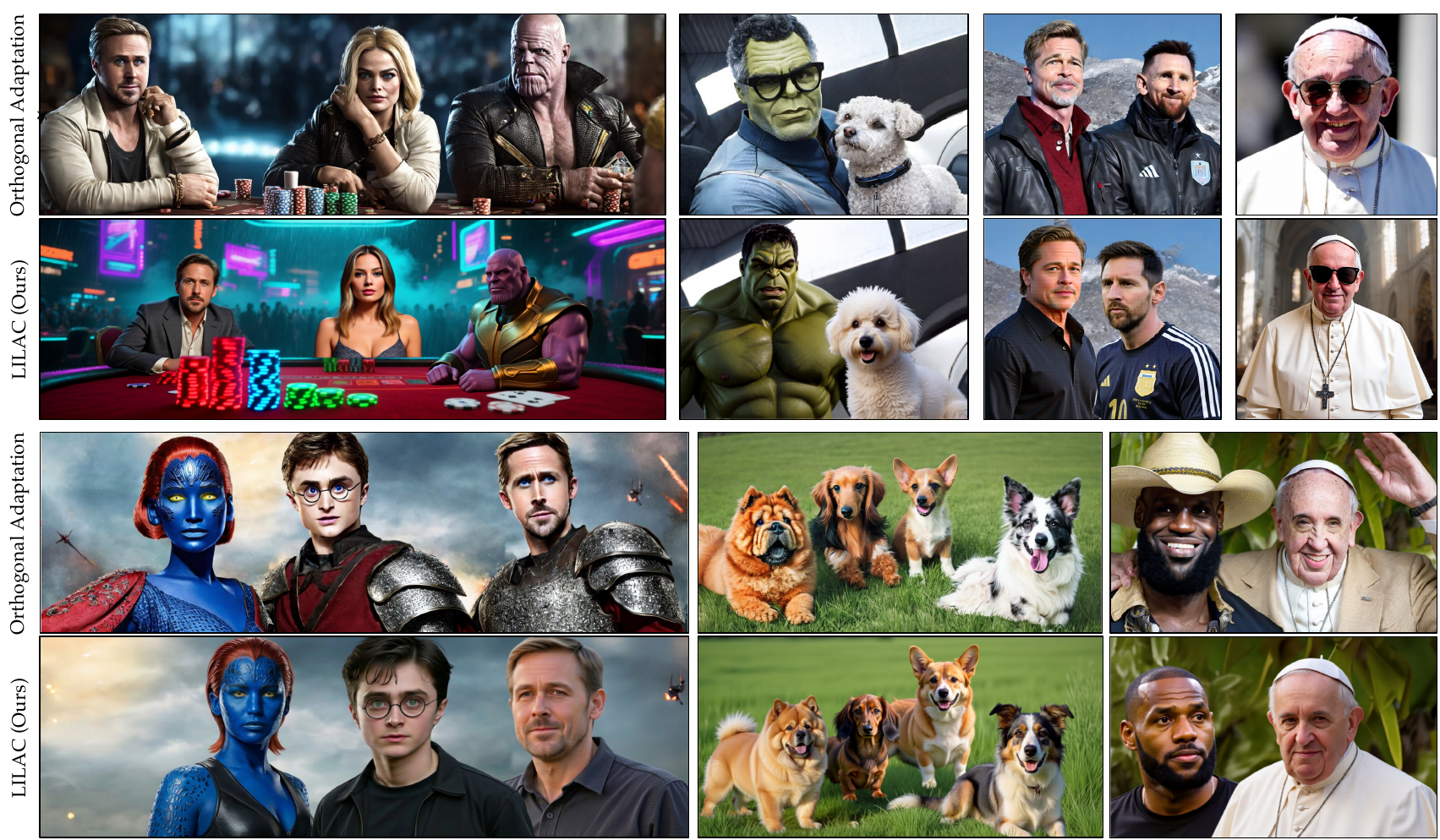}
\vspace{-0.6cm}
\caption{Qualitative multi-subject comparison on the Orthogonal Adaptation concept
bank~\cite{po2024orthogonal}. Each pair of rows shows Orthogonal
Adaptation~\cite{po2024orthogonal} (top) and \method in its scaffold configuration (bottom)
on the same prompt. The groups vary in size and combine a broad range of concepts,
including real public figures, stylized fictional characters, animals, and accessories,
placed in diverse scene contexts.}
\label{fig:qualitative}
% \vspace{-0.5cm}
\end{figure*}

\subsection{Same-Backbone Robustness of Weight Merging}
\label{sec:robustness}
% \vspace{-0.1cm}

\begin{figure}[t]
\centering
\includegraphics[width=\linewidth]{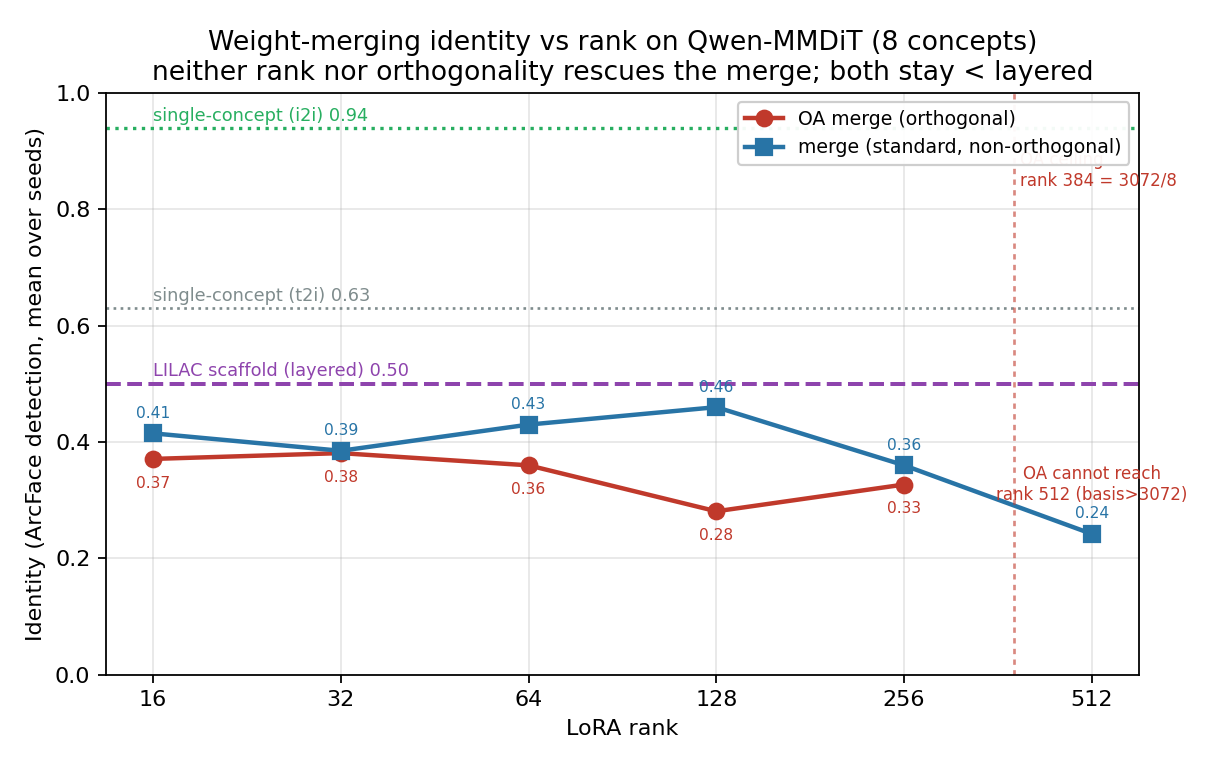}
\vspace{-0.55cm}
\caption{Same-backbone merge identity vs.\ LoRA rank on Qwen-Image-Edit ($8$ concepts,
identity averaged over seeds). Neither rank nor orthogonality rescues the merge: the
standard non-orthogonal merge matches or exceeds OA at every rank, and both stay far below
layered composition (\method scaffold). OA cannot exceed rank $384$ (basis width
$8r\le3072$).}
\label{fig:rankortho}
\vspace{-0.5cm}
\end{figure}

The value of $0.000$ reported in~\cref{tab:comparison} is a threshold effect rather than a porting error: the raw merged
three-subject cosine averages $0.29$, just below the $0.32$ cutoff, while concept fidelity
stays high at $0.81$. We independently re-trained OA adapters with extensive hyperparameter tuning, and report the resulting scores threshold-free and
averaged over seeds without screening in \cref{fig:rankortho}. The failure is robust to the baseline's own
hyperparameters: merged identity stays flat across LoRA rank $16$--$256$
(\cref{fig:rankortho}) and merge scale $[0.25,2.0]$, and degrades further as subjects are
added; OA is also capped at rank $384$ here ($8r\le3072$). Orthogonality gives
no benefit: a standard non-orthogonal merge matches or exceeds OA at every rank
(\cref{fig:rankortho}), consistent with interference over the full token sequence rather
than disjoint spatial regions (\cref{sec:binding}). Composing the \emph{same} adapters
layer-wise instead of merging recovers identity (scaffold $0.50$ vs.\ OA merge
$0.36$), isolating composition as the operative factor.
The supplementary material reports the full sweeps.

% \vspace{-0.25cm}
\subsection{Qualitative Results}
\label{sec:qualitative}
% \vspace{-0.1cm}

\cref{fig:qualitative} shows multi-subject compositions over diverse concept types such as real public figures, fictional characters, animals, and accessories, in different scene styles. \method preserves each identity and keeps the subjects distinct, placed coherently with consistent lighting. The four-dog group and the subject wearing sunglasses show that the binding is not specific to faces, composing objects and attributes as well. Supplementary  shows further compositions across subject counts, scene styles, and interactions.

\paragraph{Scaffold vs.\ decomposition.}
On Qwen-Image-Edit~\cite{wu2025qwenimage}, the two implementations of \cref{sec:cascade} differ only in how
the per-subject layers are obtained. The scaffold configuration (\cref{fig:pipeline}), which edits subjects into a
rendered scene, preserves identity more reliably than decomposition
(\cref{fig:decomposition}, which first splits a generated scene into RGBA layers with
Qwen-Image-Layered~\cite{yin2025qwenlayered}): in \cref{tab:comparison} scaffold reaches
$0.861$ against $0.639$, at a small cost in text
and image alignment. The gap is consistent with errors introduced by the layer split, which
the downstream edit cannot recover. We therefore make scaffold our default and
report decomposition only for comparison.
% Standardized (loose gates, N=3): scaffold ID 0.861 / IA 0.738 / TA 0.711;
% decomposition ID 0.639 / IA 0.743 / TA 0.734. Optional gated retention: scaffold 22/24,
% decomposition 17/24 (add if we decide to report a usability/retention number).

\subsection{Scalability}
\label{sec:scalability}
\vspace{-0.1cm}

Because the per-concept binding of \cref{sec:binding} adds a sampling pass (never a cross
term) each time a subject is added, the per-subject quality of a composition should not
collapse as the number of subjects $N$ grows. The supplementary material reports concept
fidelity (CLIP-I), text alignment (CLIP-T), and identity (ArcFace) against
$N$ for scaffold, over random subject groups. Concept
fidelity and text alignment stay flat from $N{=}1$ to $N{=}4$ ($0.71$--$0.75$
and $0.71$--$0.76$): appearance and prompt adherence do not
degrade as subjects are added. The ArcFace rate is lower and noisier
at larger $N$ (per-group std above $0.2$) without a clean
monotone trend, which we attribute to scene crowding and smaller, more frequently occluded
faces (a face-detection effect) rather than identity blending, which the
binding removes (\cref{sec:binding}): each subject is
rendered while a single adapter is active.
% TODO (Tier 4): the ArcFace curve uses 8 random groups per N (24 scenes); for camera-ready,
% smooth it with more groups per N, or report identity cosine among *detected* faces to
% separate the detection effect from identity preservation.

\paragraph{Inference cost.}
The cost is linear in the number of subjects: $N$ diffusion passes
for the native layered cascade and $N{+}1$ for scaffold, each a single generation. On one
H200, at $30$ steps and $1536{\times}512$, a pass takes $32.9\,\text{s}$ (almost
independent of $N$), so two-, three-, and four-subject scaffold compositions take $99$,
$132$, and $165\,\text{s}$ ($3$, $4$, $5$ passes). This is the cost of avoiding weight merging: the baselines amortize a one-time merge cost
(\cref{tab:comparison}, up to roughly fifteen minutes for gradient fusion)
then generate in one pass, whereas \method repeats the pass once per
subject, which is favorable when the adapters were never merged before or few images are needed.
Reducing the pass count is left to future work.

% \subsection{Backbone agnosticism}
% \label{sec:backbones}
% The LaDe results in \cref{tab:comparison} confirm that the protocol transfers unchanged to a
% native layered backbone, with per-subject scores consistent across backbones and slightly
% higher on the layered model, which additionally yields editable per-subject RGBA layers.

\section{Limitations}
\label{sec:limitations}
% \vspace{-0.25cm}

Composing subjects sequentially under frozen conditioning has clear boundaries. Because a
frozen layer cannot be modified by a later pass, \method handles ordered occlusion naturally
but struggles with close mutual occlusion and physical contact. On
Qwen-Image-Edit, where each pass re-encodes the running
composite rather than holding intact layers, repeated passes can also accumulate small
drift in earlier subjects, which the native layered backbone avoids by keeping
prior layers fixed. This re-encoding drift, not any parameter-level interference, accounts
for the residual single-to-multi identity gap of \method on Qwen-Image-Edit in
\cref{tab:comparison}.
% Our quantitative comparison also takes the baseline numbers from
% Orthogonal Adaptation~\cite{po2024orthogonal} rather than from same-backbone re-runs, so the
% values share a common scale but are indicative across backbones; a controlled same-backbone
% re-run is a cleaner comparison we leave to future work.
Finally, the method trades sampling time for simplicity ($N$ sequential passes); reducing this cost and improving contact-rich interactions are future directions.

\section{Conclusion}
\label{sec:conclusion}
% \vspace{-0.25cm}

We presented \method, a method for multi-subject personalized generation that composes
independently trained concept adapters without merging their weights. By binding one
adapter per pass and conditioning each generation on the frozen composite of the
subjects placed so far, \method removes cross-concept interference at the parameter level structurally, needs no joint training or per-composition optimization, scales linearly, and
composes adapters trained in isolation. We validated the same
protocol on an open-source image-editing model and a native layered backbone under
the Orthogonal Adaptation protocol, showing that layered, inference-time composition is a practical alternative to weight-space fusion.
On the Orthogonal Adaptation benchmark, \method reaches an ArcFace detection rate of $0.861$ on Qwen-Image-Edit and $0.877$ on LaDe, without any weight merging.

%%%%%% Bibliography %%%%%%
{
    \small
    \bibliographystyle{ieeenat_fullname}
    \bibliography{main}
}

\clearpage
\appendix
% ======================================================================
% Supplementary body (appendix sections only). Shared source of truth.
% Used by BOTH supplementary.tex (standalone, 2-PDF WACV submission) and
% main_arxiv.tex (single-PDF arXiv version) via \input{supp_body}.
% The consumer is responsible for calling \appendix before \input-ing this,
% and (for the merged arXiv build) for resetting the figure/table counters
% and the S-numbering. Contains NO \documentclass, preamble, title, or
% bibliography.
% ======================================================================

\section{Additional Qualitative Results}
\label{sec:appendix-qual}
% \vspace{-0.25cm}

\begin{figure*}[t]
\centering
\vspace{-0.1cm}
\includegraphics[width=\linewidth]{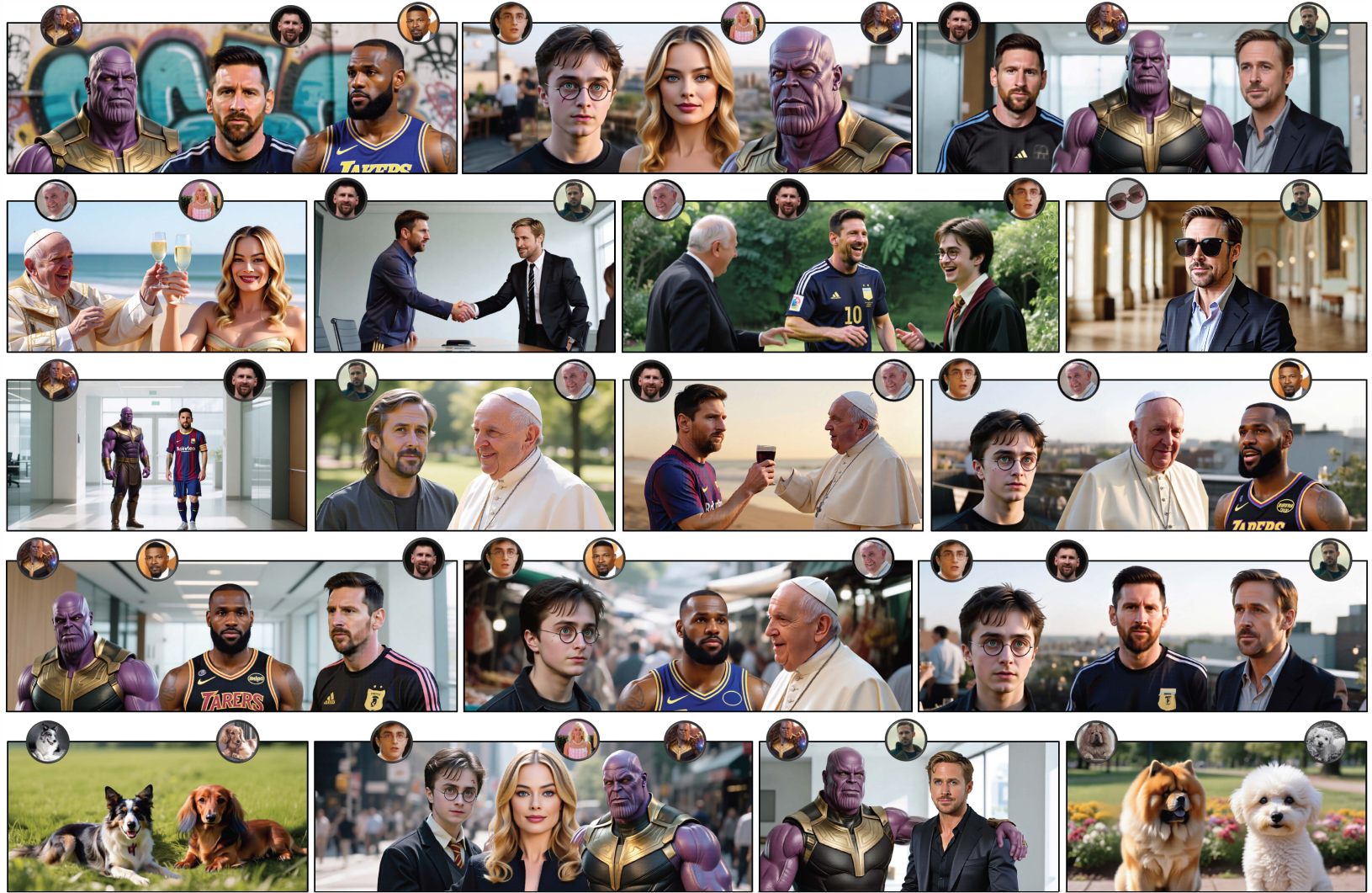}
\vspace{-0.55cm}
\caption{Additional multi-subject compositions produced by \method. Each panel shows one
generated scene; the circular thumbnails above it are the reference identities of the
composed concepts, one per subject. The examples span two to three subjects and mix real
public figures, stylized fictional characters, and animals, placed across varied scenes,
styles, and interactions (side by side, shaking hands, toasting, conversing).}
\label{fig:gallery}
% \vspace{-0.5cm}
\end{figure*}

\begin{figure*}[t]
\centering
\includegraphics[width=\linewidth]{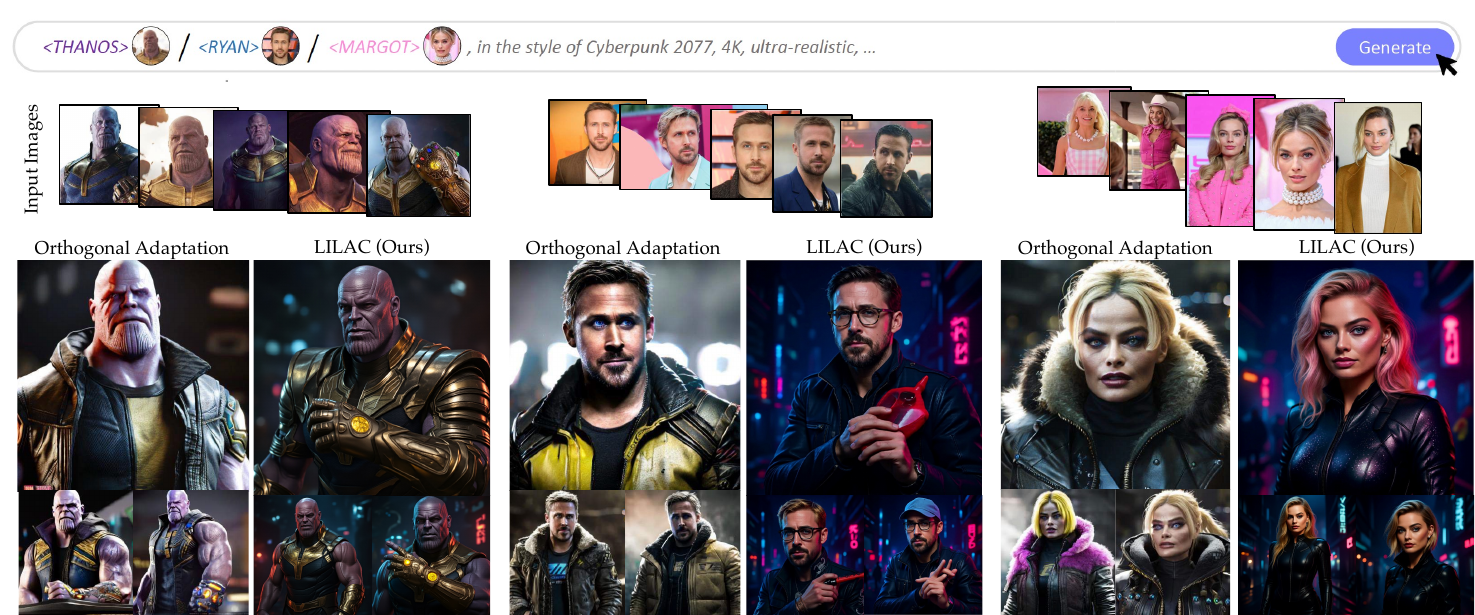}
\caption{Single-concept generations in the style of
\emph{Cyberpunk 2077} for three concepts: input references (top) and a comparison of
Orthogonal Adaptation~\cite{po2024orthogonal} with \method.}
\label{fig:appendix-qual}
\vspace{-0.5cm}
\end{figure*}

Beyond the multi-subject compositions of \cref{fig:qualitative}, we report single-concept
renderings under a pronounced artistic style, a setting that stresses identity preservation
because the style competes with each subject's appearance. For three concepts,
\cref{fig:appendix-qual} shows the reference images together with generations from
Orthogonal Adaptation~\cite{po2024orthogonal} and from \method, so that per-concept fidelity
can be compared directly.

\section{Implementation and Evaluation Details}
\label{sec:appendix-details}
% \vspace{-0.25cm}

\paragraph{Single-concept screening.}
Before composing concepts, we verify that each is captured in isolation:
we generate every concept on its own and retain only those whose single-concept identity
score clears a fixed threshold. A subject that an adapter cannot reconstruct alone cannot
be expected to survive composition, so this screen removes single-concept training
failures from the multi-subject evaluation and isolates the question we study, whether
identities are preserved when several concepts are combined, from the separate question
of whether each concept was learned well in the first place. Concepts that fail the screen
are excluded from all multi-subject results.
% TODO: state the screening threshold and list any excluded concepts (e.g. those with
% single-concept identity below threshold under text-to-image generation).

\paragraph{Implementation details.}
On the open-source backbone, each concept adapter is a rank-$64$ LoRA on the MMDiT
attention layers of Qwen-Image-Edit, trained for $1{,}000$ steps with learning rate
$10^{-4}$ and DreamBooth-style trigger-token binding on a blank source image, with the
backbone frozen throughout. At composition time the anchor is the concept with the largest
adapter delta norm $\lVert\Delta W\rVert_F$ (\cref{sec:assignment}); each subsequent pass
conditions on the frozen composite with a single adapter active, and no further
optimization is performed. For evaluation we use CLIP ViT-B/32 (OpenAI) for the alignment
metrics and InsightFace \texttt{buffalo\_l} (ArcFace) for identity, with the detection
threshold of~\cite{po2024orthogonal}.
% TODO: sampler and number of inference steps, output resolution / aspect ratio, GPU/hardware; LaDe-side adapter config.

\section{Additional Quantitative Analyses}
\label{sec:appendix-analyses}
% \vspace{-0.25cm}

\begin{figure}[t]
\centering
\includegraphics[width=1\linewidth]{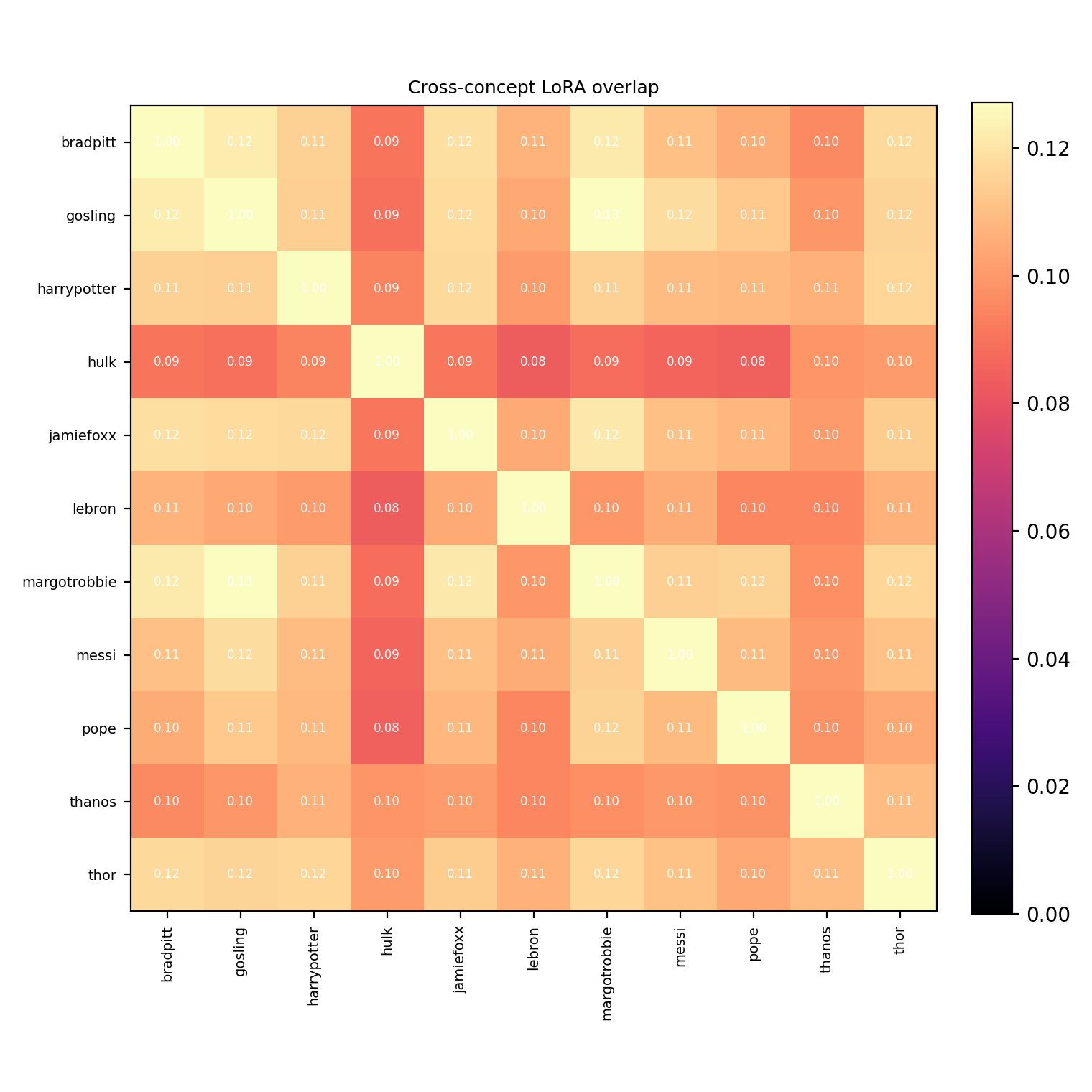}
\vspace{-0.8cm}

\caption{Cross-concept overlap between the independently trained adapters. Each cell is
the normalised overlap $\mathrm{ov}_{ij} \in [0,1]$ between the LoRA residuals of concepts
$i$ and $j$ ($1$ on the diagonal; $0$ for orthogonal residuals). Off-diagonal values
average $0.106$, $5.3\times$ the $0.020$ chance level for random adapters of the same
size: the residuals are not orthogonal. This is the interference term $\chi_{ij}$ that
weight-space fusion must suppress and that per-layer binding avoids.}
\vspace{-0.4cm}

\label{fig:crosstalk}
\end{figure}

We visualize the cross-concept overlap between the independently trained adapters in
\cref{fig:crosstalk}; it underlies the binding argument of \cref{sec:binding}.

\begin{figure}[t]
\centering
\includegraphics[width=1\linewidth]{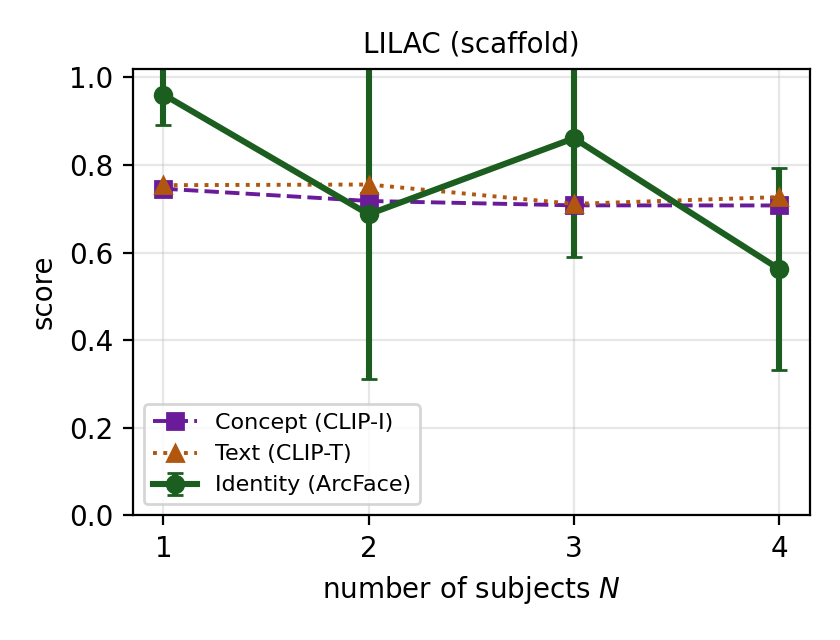}
\vspace{-0.4cm}

\caption{Per-subject quality versus the number of subjects $N$ (scaffold configuration);
$N{=}1$ is the single-concept reference, averaged over the concept bank.
Concept fidelity (CLIP-I) and text alignment (CLIP-T) stay flat as $N$ grows, while the
ArcFace detection rate is lower and noisier at larger $N$, reflecting scene crowding and
smaller faces rather than identity blending. Error bars are $\pm 1$ standard deviation,
over random subject groups for $N{\ge}2$ and over the concept bank at $N{=}1$.}

\label{fig:scalability}
\end{figure}

We report per-subject quality against the number of subjects in \cref{fig:scalability},
supporting the scalability discussion of \cref{sec:scalability}.

\section{Same-Backbone Merge Ablations}
\label{sec:appendix-ablations}
% \vspace{-0.25cm}

We detail the same-backbone sweeps summarised in \cref{sec:robustness}. Unless stated
otherwise, every configuration uses independently re-trained OA adapters on
Qwen-Image-Edit, three-subject merges over the $8$-concept person bank, and identity as
the ArcFace detection rate \emph{averaged over all seeds} (no best-of-$N$ selection) with
\emph{no} single-concept screening; ``raw'' is the raw ArcFace cosine (detection threshold
cosine $0.32$). We first verify the orthogonal basis is installed correctly on the MMDiT
modules: the cross term $\lVert B_i^{\top}B_j\rVert_F/r = 4{\times}10^{-5}$ confirms
$B_i^{\top}B_j\approx0$, so any null result below is a property of the backbone, not a
broken constraint.

\paragraph{Rank and orthogonality (\cref{tab:abl-rank}).}
We sweep the LoRA rank from $16$ to $512$ for both the orthogonal (OA) adapters and
standard non-orthogonal adapters trained with the identical recipe. Capacity does not
rescue the merge: OA identity is flat in $0.28$--$0.38$ (raw $0.27$--$0.30$) across the
whole range, far below the single-concept level. Orthogonality provides no benefit
either: the standard adapter matches or exceeds OA at every rank (e.g.\ $0.46$ vs.\ $0.28$
at rank $128$), as its frozen orthonormal up-projection trades adapter expressiveness for a
separation that does not improve the merge. OA is also structurally bounded: its shared
basis has width $(\#\text{concepts}){\times}r$, which must fit each module's output
dimension ($8r\le3072$), capping OA at rank $384$; the standard adapter runs at rank $512$
but degrades there ($0.24$) as the larger residual weakens the merge.

\begin{table}[t]
\centering
\caption{Merge identity vs.\ LoRA rank, orthogonal (OA) vs.\ standard
(non-orthogonal) adapters. Standard matches or exceeds OA at every rank; OA is
capped at rank $384$ on this backbone (basis width $8r\le3072$), where the standard
adapter still runs.}
\label{tab:abl-rank}
\setlength{\tabcolsep}{5pt}
\begin{tabular}{lcccc}
\toprule
& \multicolumn{2}{c}{Orthogonal (OA)} & \multicolumn{2}{c}{Standard} \\
\cmidrule(lr){2-3}\cmidrule(lr){4-5}
Rank & ID & raw & ID & raw \\
\midrule
16  & 0.37 & 0.27 & 0.42 & 0.31 \\
32  & 0.38 & 0.28 & 0.39 & 0.29 \\
64  & 0.36 & 0.29 & 0.43 & 0.31 \\
128 & 0.28 & 0.28 & \textbf{0.46} & 0.32 \\
256 & 0.33 & 0.30 & 0.36 & 0.28 \\
512 & ---  & ---  & 0.24 & 0.26 \\
\bottomrule
\end{tabular}
\end{table}

\paragraph{Merge scale and subject count (\cref{tab:abl-scale}).}
We sweep the per-concept merge scale $s$ (the multiplier on each $\Delta W_i$ before
summation) over $[0.25,2.0]$ for two- and three-subject merges. No scale recovers identity:
the detection rate holds near $0.40$ for $N{=}2$ and near $0.20$ for $N{=}3$ throughout, and
the grid spans the two magnitude-normalised operating points advocated to avoid an inflated
merged residual, $s{=}1/N$ and $s{=}1/\sqrt N$, neither of which helps. The single exception
($N{=}2$, $s{=}2.0$) pushes the summed residual higher: image fidelity drops while identity
rises, so this is not a real improvement. Three-subject merges lie below two-subject merges
at every scale: the more residuals are summed, the lower the identity.

\begin{table}[t]
\centering
\caption{OA merge identity vs.\ per-concept merge scale $s$, for group size $N$
(identity averaged over seeds). No scale recovers identity; $N{=}3$ is uniformly
below $N{=}2$. The scale grid includes the magnitude-normalised points
$s{=}1/N$ and $s{=}1/\sqrt N$.}
\label{tab:abl-scale}
\begin{tabular}{lcc}
\toprule
Scale $s$ & $N{=}2$ & $N{=}3$ \\
\midrule
0.25 & 0.39 & 0.14 \\
0.50 & 0.40 & 0.17 \\
0.75 & 0.40 & 0.22 \\
1.00 & 0.40 & 0.21 \\
1.50 & 0.39 & 0.31 \\
2.00 & 0.49 & 0.21 \\
\bottomrule
\end{tabular}
\end{table}

\paragraph{Composition on identical adapters (\cref{tab:abl-comp}).}
To separate the composition strategy from the adapters and the backbone, we fix the
\emph{same} re-trained OA adapters and vary only how they are combined. Merging them,
orthogonally (OA, $0.36$) or not (standard, $0.43$), keeps identity low, whereas composing
them layer-wise with our cascade recovers it: the scaffold configuration reaches $0.50$
(raw $0.35$). Because the adapters, backbone, triples, and seeds are held fixed, the gap
between merging and layering is attributable to composition alone. The single-concept
reference ($0.94$, i2i) is the ceiling every multi-subject route falls short of on this
backbone; the weaker decomposition/cascade route ($0.30$) stays in the same low range as
the merges, which is why we default to scaffold (\cref{sec:comparison}).

\begin{table}[t]
\centering
\caption{Composition on \emph{identical} adapters (only the composition strategy
varies). Layered composition beats every weight-merge, and orthogonality does not
help the merge. ID is the mean-over-seeds detection rate; ``raw'' is the raw ArcFace
cosine.}
\label{tab:abl-comp}
\setlength{\tabcolsep}{6pt}
\begin{tabular}{lcc}
\toprule
Composition (same adapters) & ID & raw \\
\midrule
Single-concept (i2i, ref.)  & 0.94 & ---  \\
\method scaffold (layered)  & \textbf{0.50} & \textbf{0.35} \\
Merge, standard (non-ortho) & 0.43 & 0.31 \\
OA merge (orthogonal)       & 0.36 & 0.29 \\
Cascade (decomposition)     & 0.30 & 0.27 \\
\bottomrule
\end{tabular}
\end{table}

\paragraph{Effect of subject count (\cref{tab:abl-n}).}
Holding the merge scale at $1.0$, we vary the number of merged subjects $N$. Concept
fidelity (CLIP-I) degrades monotonically as subjects are added ($0.81{\to}0.73{\to}0.69$ for
$N{=}2,3,4$), the expected effect of summing more residuals into one weight. The detection
rate is low throughout and noisier at larger $N$ (few random groups per $N$), so we read it
together with the fidelity trend rather than as a clean monotone curve.

\begin{table}[t]
\centering
\caption{Effect of the number of merged subjects $N$ on the OA merge (scale $1.0$).
Concept fidelity (IA) drops monotonically with $N$, while identity stays low.}
\label{tab:abl-n}
\begin{tabular}{lccc}
\toprule
$N$ & 2 & 3 & 4 \\
\midrule
Identity (ID)         & 0.35 & 0.17 & 0.28 \\
Concept fidelity (IA) & 0.81 & 0.73 & 0.69 \\
\bottomrule
\end{tabular}
\end{table}

\begin{table}[t]
\centering
\caption{Effect of subject ordering on the scaffold configuration, over a fixed set of
eight three-subject groups. Anchor-first (our default) orders subjects by descending LoRA
Frobenius norm $\lVert\Delta W\rVert_F$. Best per column in bold; higher is better.}
\label{tab:ablation_order}

\begin{tabularx}{\columnwidth}{@{}Xccc@{}}
\toprule
Ordering & Identity $\uparrow$ & Image $\uparrow$ & Text $\uparrow$ \\
\midrule
Random
& \textbf{0.850} & \textbf{0.766} & 0.687 \\

Anchor-last (ascending)
& 0.822 & 0.730 & 0.709 \\

Anchor-first (descending, ours)
& \textbf{0.850} & 0.723 & \textbf{0.714} \\
\bottomrule
\end{tabularx}
\end{table}

\section{Subject-Ordering Ablation}
\label{sec:ablation}
% \vspace{-0.25cm}

\cref{sec:assignment} orders the subjects of a composition by descending LoRA Frobenius
norm, placing the strongest-adapter concept first. To test whether this heuristic matters,
we fix a set of eight three-subject groups and compose each under three orderings: our
default (anchor-first, descending $\lVert\Delta W\rVert_F$), its reverse (anchor-last,
ascending), and a random permutation, holding every other setting identical.
\cref{tab:ablation_order} reports the outcome. The three orderings fall within a narrow band
on every metric (identity $0.82$--$0.85$, image $0.72$--$0.77$, text $0.69$--$0.71$), so the
method is robust to composition order. Within that band, anchor-first ties for the
best identity and gives the best text alignment, a random order is marginally ahead on
concept fidelity, and anchor-last is weakest on identity. The margins are
small (identity varies by under three points across orderings), so placing the strongest
adapter first is a reasonable default, consistent with treating the
order as a compositional-priority control (\cref{sec:assignment}). Absolute scores here use
a fixed eight-group subset and are not directly comparable to \cref{tab:comparison}; only
the relative ordering is meaningful.

\end{document}